\documentclass[10pt,twocolumn,letterpaper]{article}

\usepackage{wacv}
\usepackage{times}
\usepackage{epsfig}
\usepackage{color}
\usepackage{mathrsfs}  
\usepackage{amssymb,amsmath,bm}
\usepackage{multicol,multicap,graphicx}
\usepackage{subcaption}
\usepackage{wrapfig}
\usepackage{picinpar}
\usepackage{cutwin}
\usepackage{algorithm}  
\usepackage{algorithmic}
\usepackage{amsfonts}
\usepackage{amsfonts,amssymb}
\usepackage{url}
\usepackage{stmaryrd}
\usepackage{enumitem}

\usepackage[pagebackref=true,breaklinks=true,letterpaper=true,colorlinks,bookmarks=false]{hyperref}

\wacvfinalcopy 

\def\wacvPaperID{2} 

\ifwacvfinal\fi
\begin{document}

\title{Impact of ImageNet Model Selection on Domain Adaptation}

\author{Youshan Zhang \ \ \ \ \ \ \ \  Brian D.\ Davison \\
Computer Science and Engineering, Lehigh University, Bethlehem, PA, USA\\
{\tt\small \{yoz217, bdd3\}@lehigh.edu}
}

\maketitle
\ifwacvfinal\thispagestyle{empty}\fi

\begin{abstract}
Deep neural networks are widely used in image classification problems. However, little work addresses how features from different deep neural networks affect the domain adaptation problem. Existing methods often extract deep features from one ImageNet model,  without exploring other neural networks. In this paper, we investigate how different ImageNet models affect transfer accuracy on domain adaptation problems. We extract features from sixteen distinct pre-trained ImageNet models and examine the performance of twelve benchmarking methods when using the features. Extensive experimental results show that a higher accuracy ImageNet model produces better features, and leads to higher accuracy on domain adaptation problems (with a correlation coefficient of up to 0.95). We also examine the architecture of each neural network to find the best layer for feature extraction. Together, performance from our features exceeds that of the state-of-the-art in three benchmark datasets.
\end{abstract}

\vspace{-0.3cm}
\section{Introduction}
In recent years, we have witnessed the great success of deep neural networks in some standard benchmarks such as ImageNet \cite{deng2009imagenet} and CIFAR-10 \cite{krizhevsky2010convolutional}. However, in the real world, we often have a serious problem that lacks labeled data for training. It is known that training and updating of the machine learning model depends on data annotation.  Although we can get a large amount of data, few data are correctly labeled. Data annotation is a time-consuming and expensive operation. 
This brings challenges to properly train and update machine learning models.
As a result, some application areas have not been well developed due to insufficient labeled data for training.
Therefore, it is necessary to reuse existing labeled data and models for labeling new data. However, we often encounter the problem of domain shift if we train on one dataset and test  on another.

Existing work only addresses how ImageNet models affect the general classification problem~\cite{kornblith2019better}. However, no work considers how different ImageNet models affect domain adaptation.

\begin{figure}[t]
\centering
\includegraphics[width=1\columnwidth]{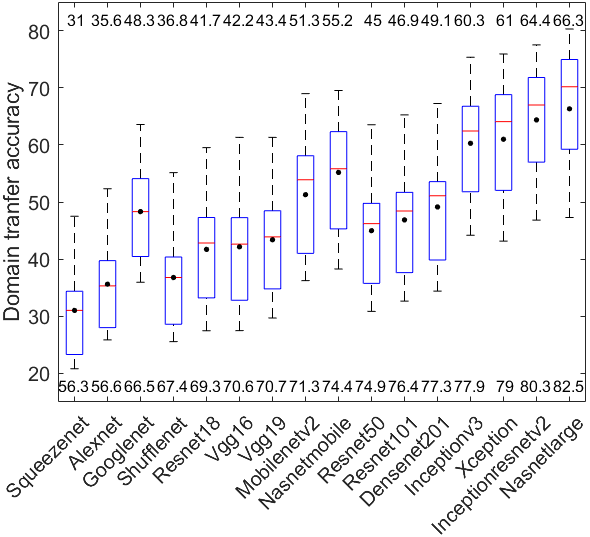}
\caption{Boxplots of domain transfer accuracy across twelve methods on each of sixteen neural networks using the Office-Home dataset (bottom is the accuracy of the ImageNet models; top is the mean domain transfer accuracy across twelve methods; black dots are mean values and red line is the median value). }
\label{fig:box}
\vspace{-0.25cm}
\end{figure}

In this paper, we report the effect of different ImageNet models on three different domain adaption datasets using each of twelve methods. We want to find how the features from these deep neural networks affect the final domain transfer accuracy. Specifically, we conduct a large-scale study of transfer learning across 16 modern convolutional neural networks for image classification on office + caltech 10, office 31 and office-home image classification datasets using two basic classifiers and ten domain adaptation methods using extracted features. Fig.~\ref{fig:box} presents boxplots of the performance of twelve methods across sixteen different neural networks using the Office-Home dataset.

This paper provides two specific contributions:
\begin{enumerate}
    \item We are the first to examine how different ImageNet models affect domain transfer accuracy, using features from sixteen distinct pre-trained neural networks on twelve methods across three benchmark datasets. The correlation of domain adaptation performance and ImageNet classification performance is high, ranging from 0.71 to 0.95, suggesting that features from a higher-performing ImageNet-trained model are more valuable than those from a lower-performing model. 
    \item We also 
    find that all
    three benchmark datasets suggest that the layer prior to the last fully connected layer is the best source. 
\end{enumerate}

\section{Background}
\subsection{Related work}
Domain adaptation has emerged as a prominent method to solve the domain shift problem. There have been efforts for both traditional \cite{gong2012geodesic,jiang2017integration,wang2018visual,zhang2019transductive} and deep learning-based \cite{vincent2008extracting,tzeng2014deep,long2015learning,ganin2016domain} methods in domain adaptation.

Traditional methods highly depend on the extracted features from raw images. Before the emergence of deep neural networks, lower-level SURF features have been widely used in domain adaptation \cite{gong2012geodesic}. However, with the development of deep neural networks, extracted features from pre-trained neural networks lead to higher performance than the use of lower-level features (Alexnet \cite{krizhevsky2012imagenet}, Decaf \cite{wang2018visual}, Resnet50 \cite{long2017deep}, Xception \cite{zhang2019transductive}, etc.).  
Distribution alignment, feature selection, and subspace learning are three frequently used methods in  traditional domain adaptation. There are also many methods that address the different kinds of distribution alignment, from marginal distribution alignment \cite{pan2011domain,dorri2012adapting,long2014transfer,jiang2017integration}, to conditional distribution alignment \cite{gong2016domain,wang2018stratified} and finally joint alignment of these two distributions \cite{wang2017balanced,wang2018visual}. Feature selection methods aim to find the shared features between source and target domain \cite{blitzer2006domain,long2014transfer}.
Subspace learning includes transfer component analysis \cite{pan2011domain} in Euclidean space, and the Riemannian subspace space includes sampling geodesic flow \cite{gopalan2011domain}, geodesic flow kernel (GKF) \cite{gong2012geodesic}, and geodesic sampling on manifolds (GSM) \cite{zhang2019transductive}. However, the predicted accuracy of traditional methods is affected by the extracted features from deep neural networks. It is believed that in general, a better ImageNet model will produce better features than a lower accuracy model on ImageNet \cite{kornblith2019better}. However, there is no such work to validate this hypothesis in domain adaptation.

Recently, deep learning models have been treated as a better mechanism for feature representation in domain adaptation. There are four major types of methods in deep domain adaptation: discrepancy-based methods, adversarial discriminative models, adversarial generative models, and data reconstruction-based models. Among all of these, maximum mean discrepancy (MMD) is one of the most efficient ways to minimize the discrepancy between source and target domain \cite{tzeng2014deep,long2015learning,ghifary2015domain}. Adversarial discriminative based models aim to define a domain confusion objective to identify the domains via a domain discriminator. The Domain-Adversarial Neural Network (DANN) considers a minimax loss to integrate a gradient reversal layer to promote the discrimination of source and target domain \cite{ganin2016domain}.
The Adversarial Discriminative Domain Adaptation (ADDA) method uses an inverted label GAN loss to split the source and target domain, and features can be learned separately \cite{tzeng2017adversarial}. The adversarial generative models combine the discriminative model with generative components based on Generative Adversarial Networks (GANs) \cite{goodfellow2014generative}.  Coupled Generative Adversarial Networks \cite{liu2016coupled} consist of a series of GANs, and each of them can represent one of the domains. Data reconstruction-based methods jointly learn source label predictions and unsupervised target data reconstruction \cite{bousmalis2016domain}. 

Both traditional and deep learning-based methods more or less rely on the extracted feature from deep neural networks. However, with so many different deep neural networks, we do not know which one is the best. Therefore, it is necessary to explore how different pre-trained models affect domain transfer accuracy.  We focus exclusively on models trained on ImageNet because it is a large benchmark dataset and pre-trained models are widely available.

\begin{figure}[h]
\centering
\includegraphics[width=1\columnwidth]{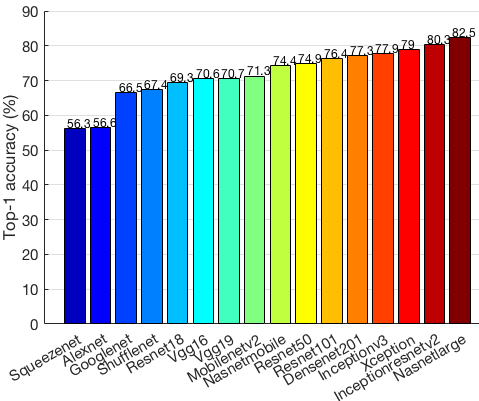}
\caption{Top-1 accuracy of the sixteen neural networks on the ImageNet task. }
\label{fig:top-1}
\vspace{-0.5cm}
\end{figure}

\subsection{ImageNet models}

There are many deep neural networks well-trained on ImageNet with differing accuracy. Kornblith et al.~\cite{kornblith2019better} explored the effects of sixteen variations of five models (Inception, Resnet, Densenet, Mobilenet, Nasnet) trained on ImageNet on general transfer learning (there is no domain shift in datasets).  In contrast, here we explore sixteen different neural network architectures from  light-weight but low performing networks to expensive, but high performing networks that have been proposed in the last decade.  Specifically, these sixteen neural networks are Squeezenet~\cite{iandola2016squeezenet}, Alexnet~\cite{krizhevsky2012imagenet}, Googlenet~\cite{szegedy2015going}, Shufflenet \cite{zhang2018shufflenet}, Resnet18~\cite{he2016deep}, Vgg16~\cite{simonyan2014very}, Vgg19~\cite{simonyan2014very}, Mobilenetv2~\cite{sandler2018mobilenetv2}, Nasnetmobile~\cite{zoph2018learning}, Resnet50~\cite{he2016deep}, Resnet101~\cite{he2016deep}, Densenet201~\cite{huang2017densely}, Inceptionv3~\cite{szegedy2016rethinking}, Xception~\cite{chollet2017xception}, Inceptionresnetv2~\cite{szegedy2017inception}, Nasnetlarge~\cite{zoph2018learning}.

\begin{figure}[h]
\centering
\includegraphics[width=1\columnwidth]{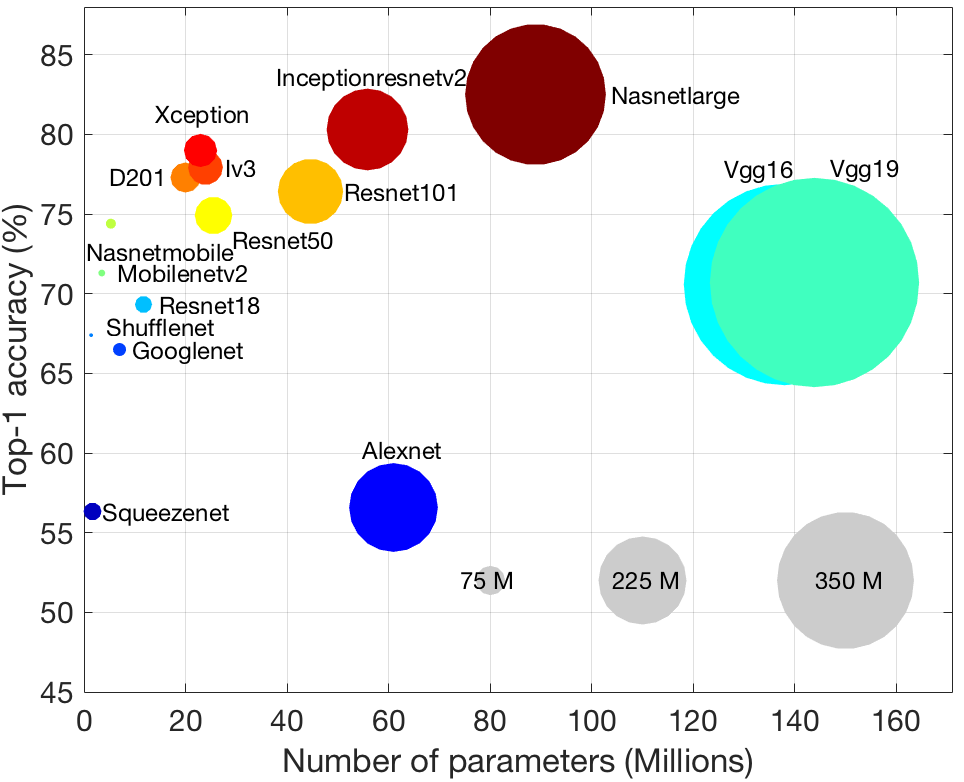}
\caption{Top-1 accuracy versus network size and parameters. (D201: Densenet201; Iv3: Inceptionv3)}
\label{fig:top_1_acc_mem}
\end{figure}

Fig.~\ref{fig:top-1} shows the top-1 classification accuracy of the sixteen neural networks on ImageNet task.  Performance ranges from 56.3\% (Squeezenet) to 82.5\% (Nasnetlarge). In  this paper, we  examine the domain adaptation feature sources according to their ImageNet accuracy. Fig.~\ref{fig:top_1_acc_mem} shows top-1 accuracy and number of parameters and network size in each  network; the gray circles show the size of memory (megabyte) and other colors represent the different models.

\subsection{Domain transfer problem}

Most previous domain adaptation methods focus only on extracted features from one neural network without exploring other networks. Resnet50 is one of the most frequently used models in this field. Kornblith et al.~\cite{kornblith2019better} pointed out that Resnet is the best source for extracting features for transfer learning. However, there is no reason to suggest this source is the best for domain adaptation, but many more methods are developed based on this source without justification. 
In our experience, extracting features from a better ImageNet model will lead to better domain adaptation performance \cite{zhang2019transductive}.
However, it is unclear whether this conclusion applies across the domain adaptation field. To address this question, we perform  extensive experiments to show the effects of features from different ImageNet models on domain transfer accuracy for domain adaptation.

\section{Methods}

\subsection{Problem and notation}

For unsupervised domain adaptation, given the source domain $\mathcal{D_S}$ data: $\mathcal{X_S}$ with its labels $\mathcal{Y_S}$ in $C$ categories and target domain $\mathcal{D_T}$ data $\mathcal{X_T}$ without any labels ($\mathcal{Y_T}$ for evaluation only). Our ultimate goal is to predict the label in the target domain with a high accuracy using the trained model from source domain.

\subsection{Feature extraction}
\begin{figure*}[h]
\centering
\includegraphics[width=2\columnwidth]{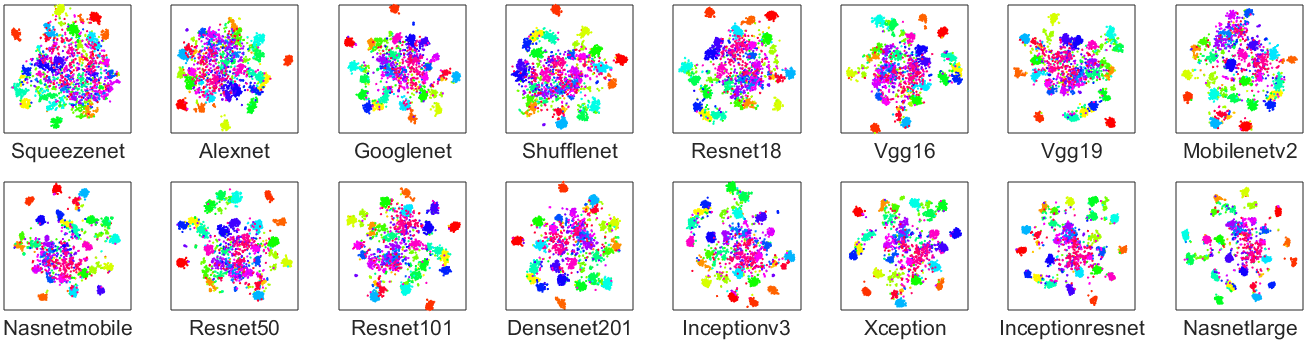}
\caption{t-SNE view of extracted features from the last fully connected layer of sixteen neural networks.  Different colors represent different classes.  The more separation of the classes in the dataset, the better the features are (Amazon domain in the Office31 dataset). }
\label{fig:tsne}
\end{figure*}

We extract the features from raw images using the above sixteen pre-trained neural networks. To get consistent numbers of features, we extract the features from the activations of the last fully connected layer~\cite{zhang2018automated,zhang2019modified}; thus the final output of one image becomes one vector $1\times 1000$. Feature extraction is implemented via two steps: (1) rescale the image into different input sizes of pre-trained neural networks; (2) extract the feature from the last fully connected layer. Fig.~\ref{fig:tsne} presents the t-SNE view of extracted features from sixteen neural networks of the Amazon domain in Office31 dataset.

\begin{figure}[h]
\centering
\includegraphics[width=1\columnwidth]{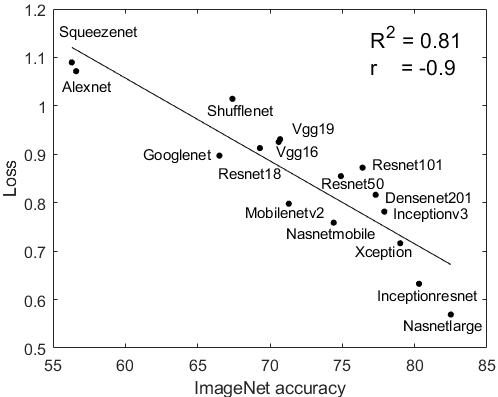}
\caption{t-SNE loss of sixteen neural networks on domain Amazon in the office31 dataset. With increase of ImageNet accuracy, the loss is reduced, representing better features.}
\label{fig:loss}
\end{figure}
\subsection{Different classifiers}

To avoid the bias that might come from a particular classification or domain adaptation method, we evaluate the performance of extracted features across twelve methods.

\begin{enumerate}[noitemsep]
\item \textbf{Support vector machines (SVM)} and \textbf{1-nearest neighbor (1NN)} are two basic classifiers that do not perform domain adaptation.  Thus, these classifiers reveal the fundamental accuracy from raw features, and serve as baselines for comparison with domain adaptation methods.
\item \textbf{Geodesic Flow Kernel (GFK)} \cite{gong2012geodesic}, which learns the ``geodesic" features from a manifold.
\item  \textbf{Geodesic sampling on manifolds (GSM)} \cite{zhang2019transductive}, which performs generalized subspace learning on manifolds.
\item  \textbf{Balanced distribution adaptation   (BDA)} \cite{wang2017balanced}, which tackles the imbalanced data and leverages marginal and conditional distribution discrepancies.
\item  \textbf{Joint distribution alignment (JDA)} \cite{long2013transfer}, which changes both marginal and conditional distribution.
\item \textbf{CORrelation Alignment (CORAL)} \cite{zhang2019transductive}, which performs second-order subspace alignment.
\item \textbf{Transfer Joint Matching (TJM)} \cite{long2014transfer}, which changes the marginal distribution by source sample selection.
\item \textbf{Joint Geometrical and Statistical Alignment (JGSA)} \cite{zhang2017joint},
which aligns marginal and conditional distributions with label propagation.
\item \textbf{Adaptation Regularization (ARTL)} \cite{long2013adaptation} learns an adaptive classifier via optimizing the structural risk function and the distribution matching between domains, and the manifold marginal distribution.
\item \textbf{Manifold Embedded Distribution Alignment  (MEDA)} \cite{wang2018visual} addresses degenerated features transformation and unevaluated distribution alignment.
\item \textbf{Modified Distribution Alignment (MDA)} \cite{zhang2019modified} is based on the MEDA model, but it removed the GFK model and replaced it with well-represented features.
\end{enumerate}

\subsection{Statistical methods}
Analysis of the domain transfer accuracy and ImageNet accuracy requires consideration of the relationship between them. We hence report the correlation coefficient score and coefficient of determination scores.

\textbf{Correlation coefficient.} We examine the strength of the correlation between ImageNet accuracy and the accuracy of the domain adaptation accuracy.
\begin{equation}\label{eq:corre}
\begin{aligned}
r(A,B)=\frac{\sum_m \sum_n (A_{mn}-\Bar{A}) (B_{mn}-\Bar{B})}{\sqrt{(\sum_m \sum_n A_{mn}-\Bar{A})^{2}} (\sum_m \sum_n B_{mn}-\Bar{B})^{2}},
\end{aligned}
\end{equation}
where $\Bar{A}$ and $\Bar{B}$ is average of vector elements. The range of the correlation is from $-1$ (strong negative) to $1$ (strong positive), while $0$ indicates there is no correlation between sub-source data and sub-target data.

\textbf{Coefficient of determination $\bf{R^2}$.} The $R^2$ statistic has proven to be a useful metric to indicate the significance of  linear regression models \cite{edwards2008r2}. The range of the $R^2$ statistic is between $[0, 1]$; the higher the $R^2$ value, the more variation is explained by the model, and the better the model fits data.
\begin{equation}\label{eq:r2}
    R^2=1-\frac{\text{Unexplained variation }}{\text{Total variation}}=1-\frac{\sum_{i=1}^N S_{residual}}{\sum_{i=1}^N S_{total}},
\end{equation}
where $S_{residual}=(y_{i} - y_{i}')^2$, $S_{total}= (y_{i} - \Bar{y})^2$, $N$ is the number of samples, $y_{i}'$ is the estimate from the regression model, $y_{i}$ is the actual value, $\Bar{y}$ is the mean value of $y$. 

By evaluating methods using the above metrics, we can determine the relationship between features from ImageNet models and domain transfer accuracy.

\vspace{-0.3cm}

\section{Results}
\subsection{Datasets}

We calculate the accuracy of each method in the image recognition problem of three datasets\footnote{Source code is available at \url{https://github.com/heaventian93/ImageNet-Models-on-Domain-Adaptation}.}. 

\textbf{Office + Caltech 10} \cite{gong2012geodesic} is a standard benchmark for domain adaptation, which consists of Office 10 and Caltech 10  datasets. It contains 2,533 images in four domains: Amazon (A), Webcam (W), DSLR (D) and Caltech (C). Amazon images are mostly from online merchants,  DSLR  and Webcam images are mostly from offices. In the experiments, C $\shortrightarrow$ A means learning knowledge from domain C and applied to domain A. We evaluate all methods and networks across twelve transfer tasks. 

\textbf{Office-31} \cite{saenko2010adapting} is another benchmark dataset for domain adaptation, and it consists of 4,110 images in 31 classes from three  domains: Amazon (A), which contains images from amazon.com, Webcam (W), and DSLR (D), both containing images that are taken by a web camera or a digital SLR camera with different settings, respectively. We evaluate all methods on all six transfer tasks A$\shortrightarrow$W, D$\shortrightarrow$W, W$\shortrightarrow$D, A$\shortrightarrow$D, D$\shortrightarrow$A, and W$\shortrightarrow$A.

\textbf{Office-Home} \cite{venkateswara2017deep} contains 15,588 images from four domains, and it has  65 categories. Specifically, Art (Ar) denotes artistic depictions for object images, Clipart (Cl) describes picture collection of clipart, Product (Pr) shows object images with a clear background and is similar to Amazon category in Office-31, and Real-World (Rw) represents object images collected with a regular camera. We also have twelve tasks in this dataset.

\begin{figure*}[h!]
\centering
\begin{subfigure}{1\textwidth}
\includegraphics[width=\linewidth]{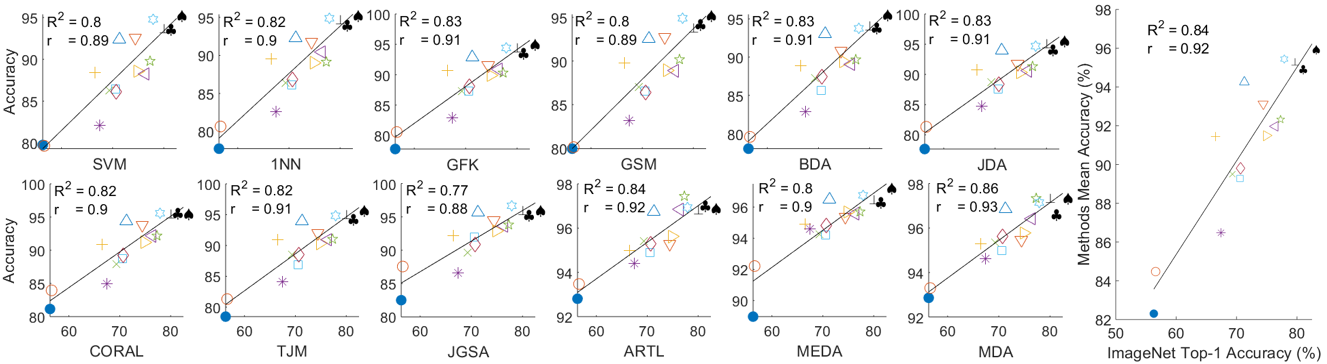}
\caption{Correlation and $R^2$ square value of Office + Caltech 10 dataset} \label{fig:ima}
\end{subfigure}\ \ \ 
\begin{subfigure}{1\textwidth}
\includegraphics[width=\linewidth]{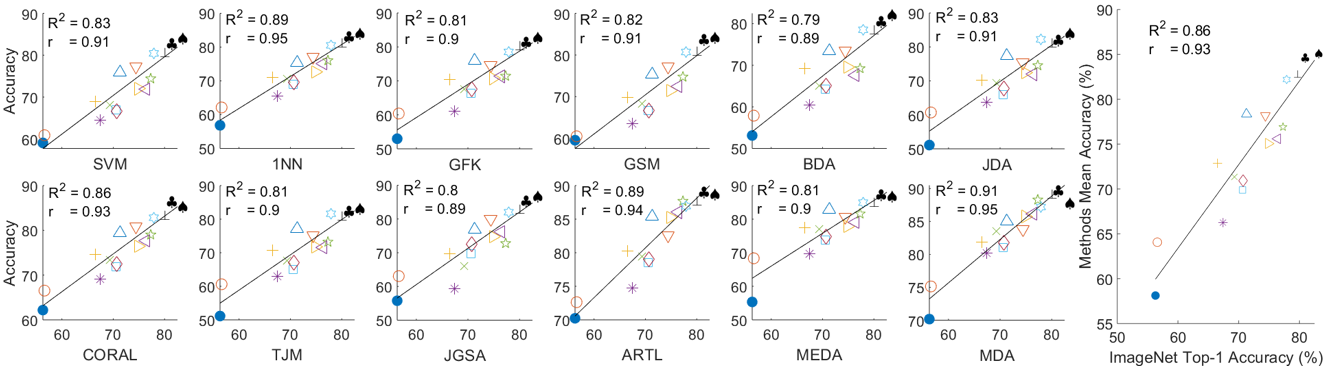}
\caption{Correlation and $R^2$ square value of Office31 dataset} \label{fig:imc}
\end{subfigure} 
\begin{subfigure}{1\textwidth}
\includegraphics[width=\linewidth]{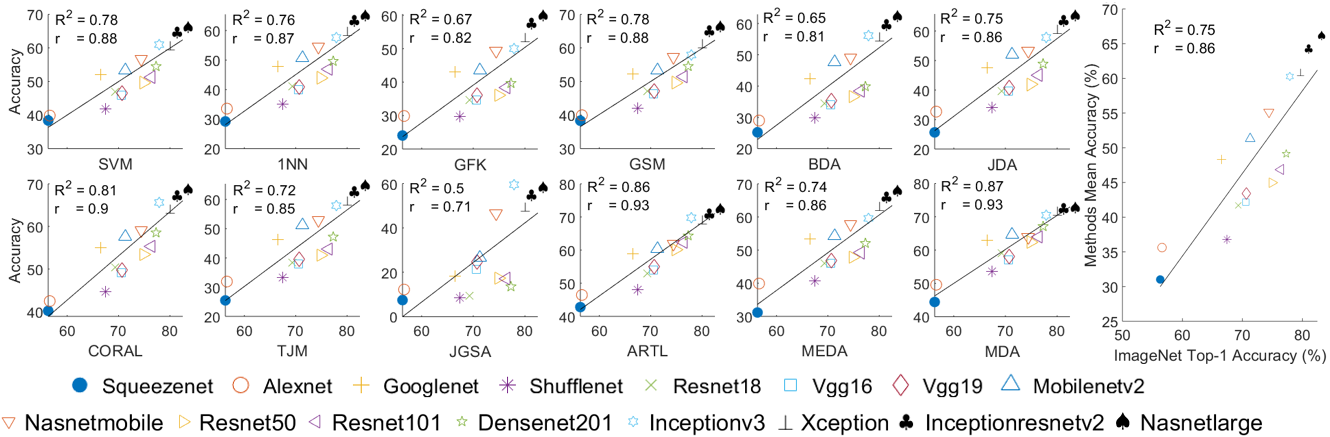}
\caption{Correlation and $R^2$ square value of Office-Home dataset} \label{fig:imd}
\end{subfigure}
\caption{The relationship between ImageNet models and three bookmarking domain adaptation datasets with 12 methods. In each subfigure, the left is the relationship between ImageNet and the domain transfer accuracy across twelve methods, and the right is the average performance of twelve methods.} \label{fig:rela}
\vspace{-0.3cm}
\end{figure*}

To get a consistent number of features, we extract features from the last fully connected layer. Fig.~\ref{fig:tsne} represents the t-SNE view of extracted features from the sixteen neural networks.  With the increasing of ImageNet classification accuracy, the separation of  features is also improved, which indicates the features are better (from the mixed colors of Squeezenet to the more clearly separated Nasnetlarge model). Also, t-SNE projection loss illustrates the goodness of features.  The loss function of the t-SNE method is Kullback-Leibler divergence, which measures the difference between similarities of points in the original space and those in the projected distribution \cite{maaten2008visualizing}. Thus if the features are well separated in the original higher dimensional space, then in a successful mapping to a low-dimensional space they will also be well separated; the more similar the two distributions, the lower the loss. Typically, lower losses correspond to better features. As shown in Fig.~\ref{fig:loss}, the correlation between the loss and different neural networks is -0.9, and $R^2$ is .81, which suggests a strong relationship.


\begin{table}[h!]
\small
\begin{center}
\captionsetup{font=small}
\caption{Feature extraction time (Seconds), number of parameters (Millions), and network size (Megabytes) for each source on Office + Caltech-10 datasets.}
\setlength{\tabcolsep}{+1.8mm}{
\begin{tabular}{rlllllll}
\hline \label{tab:time}
Task & Net size & Parameters &   Time  \\
\hline
Squeezenet~\cite{iandola2016squeezenet} & 46 & 1.24 & 13.3\\
Alexnet~\cite{krizhevsky2012imagenet} & 227 & 61& 13.9 \\
Googlenet~\cite{szegedy2015going}  & 27  &7 & 15.9 \\
Shufflenet~\cite{zhang2018shufflenet}  &6.3  & 1.4& 17.0  \\
Resnet18~\cite{he2016deep}   & 44 & 11.7& 14.8   \\
Vgg16~\cite{simonyan2014very}  & 515 &138 &  33.6   \\
Vgg19~\cite{simonyan2014very}  &535 & 144 & 37.1   \\
Mobilenetv2~\cite{sandler2018mobilenetv2} & 13 & 3.5 & 21.4  \\
Nasnetmobile~\cite{zoph2018learning} &20 & 5.3 & 39.3   \\
Resnet50~\cite{he2016deep}  & 96 & 25.6&  22.7   \\
Resnet101~\cite{he2016deep} & 167  & 44.6&  26.7   \\
Densenet201~\cite{huang2017densely} & 77 &  20 & 61.8   \\
Inceptionv3~\cite{szegedy2016rethinking}  & 89 & 23.9 & 28.2 \\
Xception~\cite{chollet2017xception}  &85  & 22.9 & 48.1\\
Inceptionresnetv2~\cite{szegedy2017inception}  & 209 &55.9 &  54.1   \\
Nasnetlarge~\cite{zoph2018learning}  & 360 & 88.9 & 141.2  \\
\hline
\end{tabular}}
\end{center}
\vspace{-.1in}
\end{table}

Table~\ref{tab:time} shows the size and number of parameters of different neural networks and feature extraction time\footnote{Features are extracted using a Geforce 1080 Ti.}.
We find an interesting phenomenon that some networks with larger size and more parameters use less time (e.g., Resnet101), which implies that the size of neural networks are not the only factor affecting feature extraction time. The correlation between extraction time and the network size and the number of parameters are 0.38 and 0.35, respectively, which further reflects the limits of the effects of network size and number of parameters on extraction time.

\begin{table*}[h]
\footnotesize
\begin{center}
\caption{Accuracy (\%) on Office + Caltech-10 datasets\vspace{-.1cm}}
\setlength{\tabcolsep}{+1.5mm}{
\begin{tabular}{|r|c|c|c|c|c|c|c|c|c|c|c|c||c|}
\hline \label{tab:OC+10}
Task & C $\shortrightarrow$ A &  C $\shortrightarrow$ W & C $\shortrightarrow$ D & A $\shortrightarrow$ C & A $\shortrightarrow$ W & A $\shortrightarrow$ D & W $\shortrightarrow$ C & W $\shortrightarrow$ A & W $\shortrightarrow$ D & D $\shortrightarrow$ C & D $\shortrightarrow$ A & D $\shortrightarrow$ W & \textbf{Average}\\
\hline
\textbf{SVM} & 94.7&   97.3&   99.4&   93.3&   90.5&   92.4&   93.9&   95.4&  \textbf{100} &94.2&   94.4&   99.0 & 95.4\\
\textbf{1NN}	&95.7&   96.3&   95.5&   93.6&   91.5&   95.5&   93.7&   95.7&  \textbf{100} & 93.5&   94.8&   98.3&   95.3\\
\textbf{GFK}~\cite{gong2012geodesic}&	94.8&   96.6&   94.9&   92.4&   92.5&   94.9&   93.6&   95.2&  \textbf{100} & 94.2&   94.4&   98.3&   95.2\\
\textbf{GSM}~\cite{zhang2019transductive}	&95.6&   96.3&   98.1&   93.9&   90.2&   93.0 &93.9 &   95.5&  \textbf{100} &  94.4&   94.4&   99.0 & 95.4\\
\textbf{BDA}~\cite{wang2017balanced}	&95.7&   95.6&   96.8&   92.8&   96.6&   94.9&   93.5&   95.8&  \textbf{100} &  93.3&   95.8&  96.3&   95.6\\
\textbf{JDA}~\cite{long2013transfer}&95.3&   96.3&   96.8&   93.9&   95.9&   95.5&   93.5&   95.7&  \textbf{100} & 93.3&   95.5&   96.9&   95.7\\
\textbf{CORAL}~\cite{sun2017correlation}	& 95.6&   96.3&   98.1&   95.2&   89.8&   94.3&   93.9&   95.7&  \textbf{100} &  94.0 & 96.2&   98.6&   95.6\\
\textbf{TJM}~\cite{long2014transfer}&95.7&   96.6&   95.5&   93.2&   95.9&   97.5&   93.4&   95.7&  \textbf{100} & 93.5&   95.6&   96.9&   95.8\\
\textbf{JGSA}~\cite{zhang2017joint}	&95.2&   97.6&   96.8&   95.2&   93.2&   95.5&   94.6&   95.2&  \textbf{100} & 94.9 &   96.1&   99.3&   96.1\\
\textbf{ARTL~}\cite{long2013adaptation}	&95.7&   97.6&   97.5&   94.6&   98.6&  \textbf{100} & 94.6&   96.1&  \textbf{100} & 93.5 &   95.8&   99.3&   96.9\\
\textbf{MEDA}~\cite{wang2018visual}	& \textbf{96.0} & \textbf{99.3} &   98.1&   94.2&   99.0 &100 & 94.6&   96.5 &  \textbf{100} &  94.1&   96.1&   99.3&   97.3\\
\textbf{MDA}~\cite{zhang2019modified}& \textbf{96.0} & \textbf{99.3} &   \textbf{99.4} &   \textbf{94.2} &   99.0 & \textbf{100} & \textbf{94.6} &   \textbf{96.5} &  \textbf{100} &  \textbf{94.2}&   \textbf{96.1}&   99.3 &   \textbf{97.4}\\
\hline
\hline
DAN~\cite{long2015learning} 	&	92.0&	90.6&	89.3	&	84.1&	91.8&	91.7&	81.2&	92.1&		\textbf{100}	&	80.3	&	90.0 &	98.5&		90.1\\
DDC~\cite{tzeng2014deep}	&	91.9&	85.4&	88.8&		85.0&	86.1&	89.0&	78.0&	83.8&		\textbf{100}	&	79.0	&	87.1&	97.7	&	86.1\\
DCORAL~\cite{sun2016deep} 	&	89.8&	97.3 &	91.0	&	91.9 &	\textbf{100}	& 90.5 &	83.7&	81.5&		90.1	& 88.6	&	80.1&	92.3&		89.7\\
RTN~\cite{long2016unsupervised}  &93.7 & 96.9 &94.2 &88.1 &95.2 & 95.5& 86.6& 92.5& \textbf{100} & 84.6& 93.8 & 99.2 &93.4 \\
MDDA~\cite{rahman2019minimum}  &93.6 & 95.2 &93.4 &89.1 &95.7 & 96.6& 86.5&94.8 & \textbf{100} & 84.7& 94.7 & \textbf{99.4} & 93.6\\
 \hline
\end{tabular}}
\end{center}
\vspace{-0.3cm}
\end{table*}

\begin{table*}[h]
\begin{center}
\footnotesize
\caption{Accuracy (\%) on Office-Home datasets\vspace{-.1cm}}
\setlength{\tabcolsep}{+1.0mm}{
\begin{tabular}{|r|c|c|c|c|c|c|c|c|c|c|c|c||c|}
\hline \label{tab:OH}
Task & Ar $\shortrightarrow$ Cl &  Ar $\shortrightarrow$ Pr & Ar $\shortrightarrow$ Rw & Cl $\shortrightarrow$ Ar & Cl $\shortrightarrow$ Pr & Cl $\shortrightarrow$ Rw & Pr $\shortrightarrow$ Ar & Pr $\shortrightarrow$ Cl & Pr $\shortrightarrow$ Rw & Rw $\shortrightarrow$ Ar & Rw $\shortrightarrow$ Cl & Rw $\shortrightarrow$ Pr & \textbf{Average}\\
\hline
\textbf{SVM} & 47.8	&76.1	&79.2&	61.7	&70.2&	69.5	&64.4&	48.7&	79.5&	70.6&	49.1&	82.1	&66.6 \\
\textbf{1NN}	& 46.4&	71.7&	77&	63.9&	69.6&	70.4&	65.5&	46.8&	76.0 &	71.4&	48.5&	78.7&	65.5\\
\textbf{GFK}~\cite{gong2012geodesic}&	39.6&	66.0	&72.5&	55.7&	66.4&	64.0 &	58.4&	42.5&	73.3&	66.0	&44.1&	76.1&	60.4\\
\textbf{GSM}~\cite{zhang2019transductive}	& 47.6&	76.4&	79.5&	62.2&	69.7&	69.2&	65.1&	49.5&	79.8&	71.0 &	49.6&	82.1&	66.8\\
\textbf{BDA}~\cite{wang2017balanced}	& 43.3&	69.8&	74.1&	58.7&	66.3&	67.7&	60.6&	46.3&	75.3&	67.3&	48.7&	77.0	&62.9\\
\textbf{JDA}~\cite{long2013transfer}& 47.4&	72.8&	76.1&	60.7&	68.6&	70.5&	66.0 &	49.1&	76.4&	69.6&	52.5&	79.7&	65.8 \\
\textbf{CORAL}~\cite{sun2017correlation}	& 48.0 &	78.7&	80.9&	65.7&	74.7&	75.5&	68.4&	49.8&	80.7&	73.0	&50.1&	82.4&	69.0\\
\textbf{TJM}~\cite{long2014transfer}& 47.6&	72.3&	76.1&	60.7&	68.6&	71.1&	64.0 &	49.0 &	75.9&	68.6&	51.2&	79.2&	65.4\\
\textbf{JGSA}~\cite{zhang2017joint}&42.9&	69.5&	71.2&	50.1&	63.0	&63.3&	55.6&	42.6&	71.8&	60.8&	42.1&	74.6&	59.0 \\
\textbf{ARTL}~\cite{long2013adaptation}& 53.5	&80.2&	81.6&	71.5&	79.9&	78.3&	\textbf{73.1} &	56.1&	\textbf{82.9}&	\textbf{75.9}	& \textbf{57.1} &	83.7&	72.8\\
\textbf{MEDA}~\cite{wang2018visual}& 48.5	&74.5&	78.8&	64.8&	76.1&	75.2&	67.4&	49.1&	79.7&	72.2&	51.7&	81.5	&68.3\\
\textbf{MDA}~\cite{zhang2019modified}& \textbf{54.8} &	\textbf{81.2} &	\textbf{82.3} &	\textbf{71.9} &	\textbf{82.9} &	\textbf{81.4} &	71.1&	\textbf{53.8} &	82.8&	75.5&	55.3&	\textbf{86.2} & 	\textbf{73.3}\\
\hline
\hline
DCORAL~\cite{sun2016deep}   & 32.2 &40.5 &54.5 & 31.5 & 45.8 &47.3 &30.0 &32.3 &55.3 & 44.7 & 42.8 &59.4 &42.8 \\
RTN~\cite{long2016unsupervised} &31.3 &40.2  & 54.6 &32.5 &46.6 &48.3 &28.2 &32.9 &56.4 &45.5 &44.8 &61.3 &43.5\\
DAH~\cite{venkateswara2017deep}  & 31.6 &40.8 &51.7 &34.7 &51.9 &52.8 &29.9 &39.6 &60.7 &45.0 &45.1 &62.5 &45.5\\
MDDA~\cite{rahman2019minimum} & 35.2 &44.4 &57.2 &36.8 & 52.5 &53.7 &34.8 &37.2 &62.2 &50.0 &46.3 &66.1 &48.0\\
DAN~\cite{long2015learning} 	& 43.6	& 57.0& 	67.9& 	45.8& 	56.5& 	60.4& 	44.0& 	43.6& 	67.7& 	63.1& 	51.5& 	74.3& 	56.3\\
DANN~\cite{ghifary2014domain} 	& 45.6	& 59.3& 	70.1& 	47.0& 	58.5& 	60.9& 	46.1& 	43.7& 	68.5& 	63.2& 	51.8& 	76.8& 	57.6\\
JAN~\cite{long2017deep}	& 45.9& 	61.2& 	68.9& 	50.4& 	59.7& 	61.0& 	45.8& 	43.4& 	70.3& 	63.9& 	52.4& 	76.8& 	58.3\\
CDAN-RM~\cite{long2018conditional} 	& 49.2& 	64.8& 	72.9& 	53.8& 	62.4& 	62.9& 	49.8& 	48.8& 	71.5& 	65.8& 	56.4& 	79.2& 	61.5\\
CDAN-M~\cite{long2018conditional} 	& 50.6& 	65.9& 	73.4& 	55.7& 	62.7& 	64.2& 	51.8& 	49.1& 	74.5& 	68.2& 	56.9& 	80.7& 	62.8\\
\hline
\end{tabular}}
\end{center}
\vspace{-0.6cm}
\end{table*}

\subsection{ImageNet and domain transfer accuracy}
\label{sec:imagenet}
In this setting, different neural networks are only used to extract features; we do not re-train the neural network since we want to explore purely how the different deep ImageNet models affect domain transfer accuracy. 

We examine the sixteen deep neural networks in ImageNet, and top-1 target domain accuracy ranges from 56.3\% to 82.5\%. We measure the trend of domain adaptation performance across three datasets and twelve methods using correlation and $R^{2}$ statistics.  Fig.~\ref{fig:rela} presents the correlations and $R^{2}$ statistics between top-1 accuracy on ImageNet and the performance of the domain adaptation accuracy. We can make several observations: first of all, the overall domain transfer performance from three datasets is linearly correlated with the increase of the ImageNet model performance. 

Among the three datasets, office-home is most challenging because the average performance of all twelve methods are lower than 70\% and there is more domain shift in this dataset; the overall accuracy of the other two datasets are higher than 85\%, and this leads to correlation score and $R^2$ value of the office-home dataset to be lower than the other two datasets. Secondly, in the office + caltech 10 dataset, the result from each method presents a similar trend such that with the increasing of ImageNet accuracy, the transfer accuracy is also improved.  Notably, we get a different conclusion from previous work \cite{kornblith2019better}, which stated that Resnet and Densenet usually give the highest performance. 

Thirdly, we see that Nasnetlarge currently has the highest top-1 accuracy in ImageNet. We therefore expect that the features from  Nasnetlarge would have the highest performance across three datasets, and it is true that most methods follow this observation. However, we notice that the JGSA model in Office + caltech 10 and MDA model has a lower accuracy than the Inceptionresnetv2 model, which is caused by an error in the model (invalid update of the conditional and marginal distributions). Fourth, the ARTL model has a strange relative performance in the Office-Home dataset; the transfer accuracy from Squeezenet to the Densenet201 is significantly lower than Inceptionv3, Xception, Inceptionresnetv2, and Nasnetlarge. The reason is that the JGSA model does not perform well if there is a significant difference between the source and target domains. 

\subsection{Comparison with state-of-the-art results}
\vspace{-0.3cm}
Due to space limitations, we only list the highest accuracy across three datasets using the twelve representative methods along with a few other state-of-the-art methods in Tables \ref{tab:OC+10}-\ref{tab:O31} (Office + Caltech-10 and Office-Home use features from Nasnetlarge and Office 31 uses features from Inceptionresnetv2).
The overall performance across all twelve methods is higher than state-of-the-art methods, which demonstrates  the superiority of the extracted features.  However, the classification results of twelve methods are compromised in some tasks (e.g., W $\shortrightarrow$ A and D $\shortrightarrow$ A in Office 31 datasets), which is likely caused by the differences in tasks, and we cannot guarantee  top features are best in all tasks but overall performance is significantly better.

\subsection{Which is the best layer for feature extraction?}
\vspace{-0.05cm}

We extract the features from the last fully connected layer
which corresponds to a feature size of  $1 \times 1000$. However, we do not know which layer is the best one for feature extraction in the domain adaptation problem. In this section, we give an experimental suggestion to choose the best layer for feature extraction. In Tab.~\ref{tab:OC_layer}, we show the results of the last four layers (as other layers often have an extremely large number of features).
The output and softmax layers have the same accuracy since the output just changes the probability of the softmax layer to a real class. In addition, we find that the last fully connected layer (LFC) is not the best layer for feature extraction. Instead, the layer prior to the last fully connected layer (P\_LFC) has the highest performance. The average improvement of P\_LFC layer over the LFC layer across sixteen neural networks for each of the datasets are 0.2\%,1.1\%, and 1.5\%, respectively.

\begin{table}[h!]
\footnotesize
\begin{center}
\caption{Accuracy (\%) on Office 31 datasets}
\vspace{-0.3cm}
\setlength{\tabcolsep}{+0.6mm}{
\begin{tabular}{rcccccccc|c|c|c|c|c|c|c|c|}
\hline \label{tab:O31}
Task & A $\shortrightarrow$ W &  A $\shortrightarrow$ D & W $\shortrightarrow$ A & W $\shortrightarrow$ D & D $\shortrightarrow$ A & D $\shortrightarrow$ W  & \textbf{Average}\\
\hline
\textbf{SVM} &  81.5   &80.9 &  73.4 &  96.6  & 70.6   &95.1  & 83.0\\
\textbf{1NN}	& 80.3 &  81.1  & 71.8 &  99.0 &  71.3  & 96.4  & 83.3\\
\textbf{GFK}~\cite{gong2012geodesic}&	78.1  & 78.5 &  71.7  & 98.0  & 68.9  & 95.2 &  81.7\\
\textbf{GSM}~\cite{zhang2019transductive}	& 84.8  & 82.7  & 73.5   &96.6  & 70.9  & 95.0  & 83.9\\
\textbf{BDA}~\cite{wang2017balanced}	& 77.0   &79.3 &  70.3  & 97.0  & 68.0 &  93.2 &  80.8\\
\textbf{JDA}~\cite{long2013transfer}& 79.1 &  79.7  & 72.9 &  97.4  & 71.0  & 94.2 &  82.4\\
\textbf{CORAL}~\cite{sun2017correlation}	& 88.9 &  87.6 &  74.7 &  99.2 &  73.0  & 96.7 &  86.7 \\
\textbf{TJM}~\cite{long2014transfer}& 79.1  & 81.1  & 72.9  & 96.6  & 71.2  & 94.6  & 82.6\\
\textbf{JGSA}~\cite{zhang2017joint}	&81.1  & 84.3  & 76.5  & 99.0 &  75.8  & 97.2  & 85.7\\
\textbf{ARTL}~\cite{long2013adaptation}	&92.5  & 91.8  & 76.9  & 99.6  & 77.1 &  97.5 &  89.2\\
\textbf{MEDA}~\cite{wang2018visual}	& 90.8 &  91.4 &  74.6  & 97.2 &  75.4  & 96.0 &  87.6\\
\textbf{MDA}~\cite{zhang2019modified}& \textbf{94.0}  & \textbf{92.6}  & 77.6 &  99.2  & 78.7  & 96.9  & \textbf{89.8}\\
\hline
\hline
DAN~\cite{long2015learning}	&	80.5 &	78.6 &	62.8	&99.6 &	63.6 &	97.1 &	80.4 \\
RTN~\cite{long2016unsupervised}&	84.5 &	77.5 &	64.8 &	99.4 &	66.2 &	96.8 &	81.6 \\
DANN~\cite{ghifary2014domain}	&82.0 &	79.7 &	67.4 &	99.1 &	68.2 &	96.8 &	81.6\\
ADDA~\cite{tzeng2017adversarial}	&86.2	&77.8  &68.9	&98.4 &	69.5 &	96.2	& 82.9\\
CAN~\cite{zhang2018collaborative} &	81.5 &	65.9 &	\textbf{98.2}	&85.5 &	\textbf{99.7} &	63.4 &	82.4\\
JDDA~\cite{chen2018joint}	&82.6	&79.8 &	66.7 &	99.7	&57.4	& 95.2 &80.2\\
JAN~\cite{long2017deep}&	85.4 &	84.7	&70.0 &	\textbf{99.8}	&68.6 &	\textbf{97.4} &	84.3\\
GCAN~\cite{ma2019gcan} & 82.7 & 76.4 & 62.6 & \textbf{99.8} & 64.9 & 97.1 & 80.6\\ 
\hline
\end{tabular}}
\end{center}
\vspace{-0.6cm}
\end{table}

\subsection{How to choose the neural network to improve the domain transfer accuracy?}
\vspace{-0.05cm}

Based on the above results, we suggest extracting features from the layer which is right before the last fully connected layer. The features in this layer are not only well represented but also use less memory. Moreover, although the Nasnetlarge feature has higher accuracy among most tasks, the Inceptionresnetv2 features can sometimes achieve a better or similar result compared to Nasnetlarge, e.g., the P\_LFC performance on Tab.~\ref{tab:OC_layer}, and the Inceptionresnetv2 model is substantially smaller and runs significantly faster than the Nasnetlarge model. Therefore, we recommend choosing one of these two models for feature selection.

\begin{table}[h!]
\small
\begin{center}
\captionsetup{font=small}
\caption{Average accuracy (\%) of layer selection  on Office-Home datasets with MDA method~\cite{zhang2019modified}}
\vspace{-0.3cm}
\setlength{\tabcolsep}{+1.8mm}{
\begin{tabular}{rlllllll}
\hline \label{tab:OC_layer}
Task & Output &  Softmax & LFC & P\_LFC\\
\hline
Squeezenet~\cite{iandola2016squeezenet} &    42.0  & 42.0 & \textbf{44.4} & -\\
Alexnet~\cite{krizhevsky2012imagenet} & 43.0 & 43.0 & 49.6 & \textbf{50.4} \\
Googlenet~\cite{szegedy2015going}  & 53.0  & 53.0   &62.9  & \textbf{64.2}\\
Shufflenet~\cite{zhang2018shufflenet}  &45.9  &45.9   &53.5 &  \textbf{54.7}\\
Resnet18~\cite{he2016deep}   & 49.5 & 49.5   & 59.2 & \textbf{62.0} \\
Vgg16~\cite{simonyan2014very}  &47.8 & 47.8   & 57.1 & \textbf{58.3} \\
Vgg19~\cite{simonyan2014very}  &48.4 & 48.4  &58.0 & \textbf{59.4}\\
Mobilenetv2~\cite{sandler2018mobilenetv2}  & 52.4 & 52.4  &52.4& \textbf{64.7}\\
Nasnetmobile~\cite{zoph2018learning}   &52.8 & 52.8  &63.8 & \textbf{64.6} \\
Resnet50~\cite{he2016deep}  & 50.0  &  50.0  & 62.4 & \textbf{62.5} \\
Resnet101~\cite{he2016deep} & 51.2 &  51.2   &63.9 & \textbf{64.7}\\
Densenet201~\cite{huang2017densely} & 54.3 &  54.3   &67.1  & \textbf{69.5}\\
Inceptionv3~\cite{szegedy2016rethinking}  & 57.4 & 57.4   &69.7 & \textbf{70.4} \\
Xception~\cite{chollet2017xception}  & 59.4  &  59.4 & 72.0  & \textbf{72.3}\\
Inceptionresnetv2~\cite{szegedy2017inception}  & 60.1 &   60.1  &72.8 & \textbf{73.8} \\
Nasnetlarge~\cite{zoph2018learning}  & 60.6 & 60.6   &73.3 & \textbf{73.6}\\
\hline
\end{tabular}}
\end{center}
\vspace{-0.8cm}
\end{table}

\section{Discussion}
\vspace{-0.1cm}

\begin{table}[t]
\small
   \caption{Improvement of SVM and MDA model based on lowest and highest ImageNet model (Squeez.: Squeezenet, NAST.: Nasnetlarge and Impro.: Improvement)}\label{tab:impro}
   \centering
\begin{tabular}{|r|l|l|l|l|l}
\hline
\bf Task & SVM\_Squeez.& SVM\_NAST.& Impro.\\
\hline
Office+Caltech-10   & 79.8 & 95.4 & 19.6\%   \\
Office-31  & 59.1 & 84.3  & 42.6\%   \\
Office-Home   & 38.4 & 66.6 &73.4\%  	\\
\hline
\end{tabular}
   \vspace{0.3cm}\\
\begin{tabular}{|r|l|l|l|l|l}
\hline
\bf Task & MDA\_Squeez.& MDA\_NAST.& Impro.\\
\hline
Office+Caltech-10   & 92.9 & 97.4 & 4.8\%   \\
Office-31  & 70.2 & 88.0  & 25.4\%   \\
Office-Home   & 44.4 & 73.3 &65.1\%  \\
\hline
\end{tabular}
\vspace{-0.3cm}
\end{table}


Before the rise of the convolution neural network, hand-crafted features were well used (e.g., SURF), and since deep features can substantially improve the performance of domain adaption, they are now widely used. However, most research has stuck with one pre-trained neural network, and researchers did not know which one will give the highest performance. In this paper, we are the first to present how different well-trained ImageNet models affect domain adapted classification. We have several novel observations.  By exploring how different ImageNet models affect domain transfer accuracy, we find a roughly linear relationship between them, which suggests that Inceptionresnetv2 and Nasnetlarge are better sources for feature extraction.  This differs from the conclusion in Kornblith et al.~\cite{kornblith2019better}, which says these two neural networks do not transfer well in general classification problems. We see improved performance because of the better alignment of the source and target distribution using features from these two networks, while the domain shift issue does not exist in the general classification problem considered by Kornblith. We also find, perhaps surprisingly, that the best layer for feature extraction is the layer before the last fully-connected layer.

Tab.~\ref{tab:impro} lists the improvement of SVM and MDA models using the Squeenezenet and Nasnetlarge neural networks for feature extraction. The improvement is non-trivial across all three datasets. Especially in the most difficult dataset, office-home, performance is notably improved 73.4\% in SVM and 65.1\% in MDA model. We hence can conclude that Nasnetlarge will be particularly useful in the case in which there is larger discrepancy between source and target domains. In addition,  overall improvement suggests that better neural network features will be important for domain transfer accuracy.

Although we explore how to choose the best neural network source and layer for feature extraction, we notice that features from a lower performance ImageNet-trained network can produce a higher transfer accuracy in some tasks. Therefore, more work is needed to consider the combination of extracted features to produce even higher performance.

\section{Conclusion}
\vspace{-0.1cm}
In this paper, we are the first to examine how features from many different ImageNet models affect domain adaptation. Extensive experiments demonstrate that a better ImageNet model will give a higher performance in transfer learning. We also find that the layer prior to the last fully connected layer is the best layer for extracting features. 

{\small
\bibliographystyle{ieee}
\bibliography{egbib}
}

\end{document}